\documentclass[]{fairmeta}

\title{
\vspace{-1pt}
\textbf{GaussianWAM}: Distilling Geometry and Semantics from 3D Gaussian Fields into World-Action Models
}

\author[2,3,1,*]{Zijian Zhang}
\author[2,3,1,*]{Yuqing Jiang}
\author[4,5, \ddagger]{Weitao Zhou}
\author[5]{Minglei Li}
\author[6]{Jinhao Zhang}
\author[7]{Yao Mu}
\author[8]{Xiaofan Li}
\author[9]{Hao Zhao}
\author[10,1,\ddagger]{Haibao Yu}

\affiliation[1]{Tuojing Intelligence}
\affiliation[2]{University of Chinese Academy of Sciences}
\affiliation[3]{Institute of Automation, Chinese Academy of Sciences}
\affiliation[4]{Tsinghua University}
\affiliation[5]{Simple AI}

\affiliation[6]{Harbin Institute of Technology (Shenzhen)}
\affiliation[7]{Shanghai Jiao Tong University}

\affiliation[8]{Zhejiang University}
\affiliation[9]{Institute for AI Industry Research (AIR), Tsinghua University}
\affiliation[10]{The University of Hong Kong}

\contribution[*]{equal contribution}
\contribution[\ddagger]{corresponding author}

\usepackage{amsmath,amsfonts,bm}
\usepackage{xcolor}

\def\eqref#1{equation~\ref{#1}}

\def\1{\bm{1}}

\DeclareMathAlphabet{\mathsfit}{\encodingdefault}{\sfdefault}{m}{sl}
\SetMathAlphabet{\mathsfit}{bold}{\encodingdefault}{\sfdefault}{bx}{n}

\usepackage{graphicx}
\usepackage[table]{xcolor}
\usepackage{colortbl}

\usepackage{booktabs}
\usepackage{tabularx}
\usepackage{array}
\usepackage{multirow}
\usepackage{diagbox}
\usepackage{hhline}
\usepackage{longtable}
\usepackage{makecell}
\usepackage{siunitx}
\usepackage{adjustbox}
\usepackage{wrapfig}
\usepackage{caption}   
\newcolumntype{C}{>{\centering\arraybackslash}X}

\usepackage{amsmath,amsfonts,amssymb}
\usepackage{bm}
\usepackage{nicefrac}

\usepackage{enumitem}
\setlist[itemize]{leftmargin=*}

\usepackage{caption}
\crefname{figure}{Fig.}{Figs.}
\crefname{table}{Tab.}{Tabs.}

\usepackage{xspace}
\usepackage{calc}
\usepackage{etoolbox}
\usepackage{pifont}
\usepackage{fancyvrb}

\usepackage{titletoc}

\titlecontents{section}
[1.5em] %
{\addvspace{-0.5pt}} %
{\bfseries\contentslabel{2.3em}} %
{\hspace*{-2.3em}\bfseries} %
{\bfseries\titlerule*[.5pc]{.}\contentspage} %
\titlecontents{subsection}
[3.8em] %
{\addvspace{-2.2pt}} %
{\contentslabel{2.3em}}
{\hspace*{-2.3em}}
{\titlerule*[.5pc]{.}\contentspage}

\abstract{
World-Action Models (WAMs) jointly learn future visual prediction and
action generation, using video dynamics as a representation-learning
signal for robotic manipulation. However, their video latents are
primarily optimized for visual prediction and are not explicitly
encouraged to preserve cross-view geometric structure or spatially
localized, object-relevant semantics.
We propose \textbf{GaussianWAM}, a training-time representation-enhancement
framework that organizes geometric and semantic supervision through a
3D Gaussian field. Given synchronized multi-view observations, frozen
geometry and vision foundation models provide depth, camera parameters,
and dense semantic features. GaussianWAM binds these heterogeneous
signals to shared Gaussian primitives and renders spatially aligned
semantic, depth, and coverage targets, which are distilled into the
current-observation representations of the WAM. All teacher models,
Gaussian components, and auxiliary prediction heads are removed after
training, leaving the original WAM inference path without additional
modules or forward computation.
On LIBERO-Plus, GaussianWAM improves FastWAM from 52.05\% to 71.29\%
and Cosmos Policy from 71.52\% to 77.30\%. Direct CLIP and VGGT
distillation already establishes a strong FastWAM baseline of 69.37\%,
while Gaussian-field unification further improves it to 71.29\%,
supporting the benefit of spatially organizing heterogeneous teacher
signals. GaussianWAM also improves performance on standard LIBERO and
shows positive transfer trends on RoboTwin and real-world manipulation.
These results suggest that training-time Gaussian distillation provides
a practical way to inject geometry- and semantics-related supervision
into WAM representations without changing their deployment architecture.
}
\metadata[Code]{\url{https://github.com/TuojingAI/GaussianWAM}}
\metadata[Project Website]{\url{https://tuojingai.github.io/GaussianWAM-project-page/}}

\definecolor{lightgray}{rgb}{0.95, 0.95, 0.95}

\definecolor{baselinecolor}{gray}{.9}

\begin{document}
\maketitle 
\section{Introduction}
\label{sec:intro}

\begin{figure}[t]
    \centering
    \includegraphics[width=\linewidth]{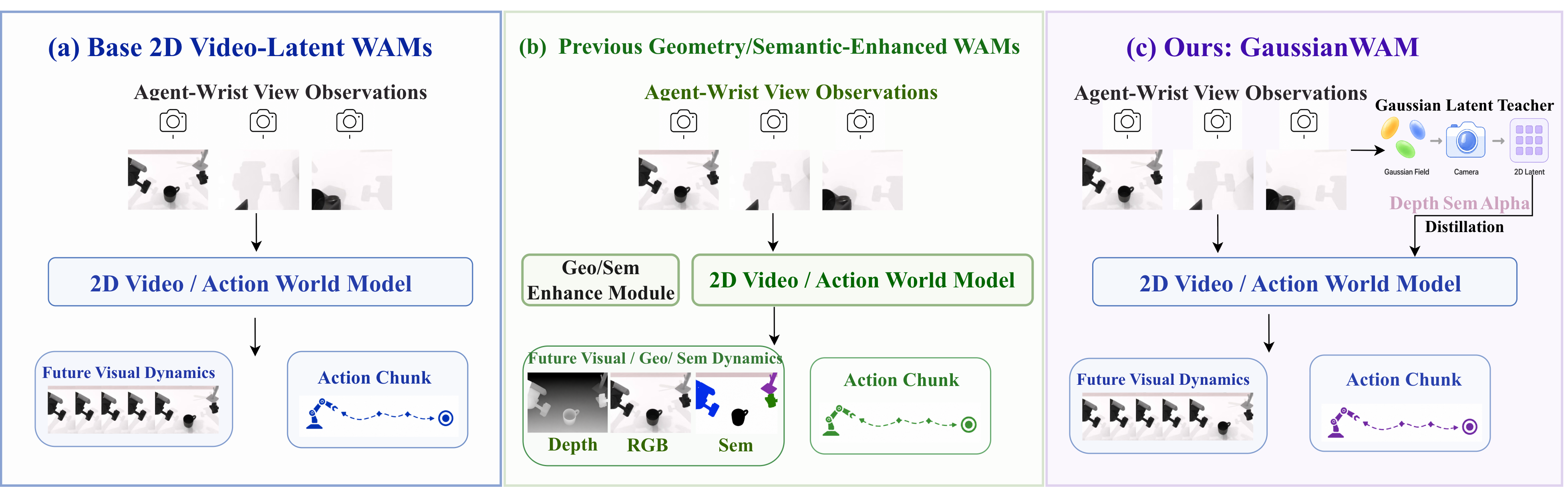}
    \caption{
    Comparison of WAM enhancement paradigms.
    (a) Conventional WAMs learn world-action representations primarily in
    2D video or latent space.
    (b) Previous geometry- or semantics-aware WAMs introduce dedicated
    spatial enhancement modules or structured representations into the
    modeling pipeline.
    (c) GaussianWAM instead uses a training-time 3D Gaussian teacher to
    inject geometry-, semantics-, and coverage-aware supervision into the
    original WAM representations, without modifying the inference backbone.
    }
    \label{fig:paradigms}
\end{figure}

Recent World-Action Models (WAMs) and video-action models have introduced a new modeling paradigm for robotic control. Unlike conventional vision-language-action (VLA) models that primarily predict robot actions directly from current observations~\citep{brohan2022rt1,kim2024openvla}, WAMs leverage the temporal dynamics and implicit physical priors learned by video generative models to jointly learn future visual dynamics and action generation~\citep{ye2026dreamzero}. As illustrated in Fig.~\ref{fig:paradigms}, this paradigm moves beyond purely reactive action prediction and uses future scene evolution as an additional representation-learning signal. In this sense, future prediction serves not only as a visual generation objective, but also as temporal supervision for action generation.

Despite this progress, the video latent representations learned by existing WAMs are primarily optimized for visual reconstruction or prediction and are not explicitly constrained to preserve cross-view geometric structure. Most WAMs model the world primarily in RGB image space or two-dimensional video latent space. While such representations can effectively capture appearance changes, motion continuity, and implicit dynamics, visual coherence does not necessarily imply geometric reliability. Robotic manipulation fundamentally takes place in three-dimensional space, where precise control requires a reliable understanding of scene geometry and spatial relationships among objects. Consequently, latent representations learned solely through RGB reconstruction or video prediction may remain geometrically under-constrained, limiting the capability of WAMs for precise spatial interaction and action generation~\citep{li2026wam4d,zhang2026mecowam,ma2026geosem,yan2026svam,zhao2026sgwam,yuan2026dreamwam,yang2026fourDwam}.

Moreover, WAM video latents are not explicitly optimized to preserve spatially localized, object-relevant semantic information. Although language instructions can condition video generation and action prediction, such conditioning does not necessarily ensure that visual tokens consistently correspond to task-relevant objects and their spatial locations. In other words, task-level language conditioning does not necessarily translate into object-relevant semantic structure in the visual latent space. Visual foundation models such as CLIP~\citep{radford2021clip}, DINO~\citep{oquab2023dinov2}, and SAM~\citep{kirillov2023sam} provide rich object- and region-level semantic priors, while geometric foundation models such as VGGT~\citep{wang2025vggt} provide depth and camera cues. Therefore, effective WAM representations for robotic manipulation should capture not only \emph{where} objects are in 3D space, but also \emph{what} the corresponding visual regions represent.

A seemingly straightforward solution is to distill geometric and semantic knowledge from multiple pretrained teachers into the WAM backbone, or to explicitly extend WAMs toward 3D/4D world modeling~\citep{guo2026xwam,li2026wam4d}. However, both directions have important limitations. Independent geometric and semantic teachers typically produce heterogeneous features in different representation spaces, viewpoints, and confidence regimes. Applying separate feature losses therefore treats geometry and semantics as disconnected supervision signals, without a shared spatial carrier that consistently associates semantic information with the same physical 3D structures across views. On the other hand, explicit 3D/4D world models often require additional 3D encoders, decoders, rendering modules, geometric annotations, or future 3D rollouts, increasing both training complexity and inference cost. These limitations raise two key questions: \emph{How can heterogeneous geometric and semantic knowledge be unified into a spatially coherent representation for WAMs? And can such structured supervision be introduced without sacrificing the inference efficiency of the original policy?}

To address these challenges, we propose \textbf{GaussianWAM}, a \textbf{Gaussian Field Enhancement framework for World-Action Models}. We use a 3D Gaussian field as a unified spatial carrier that organizes geometric and semantic supervision from different foundation models within the same set of Gaussian primitives. Specifically, multi-view geometric cues are lifted into 3D Gaussian primitives, while visual-semantic features are associated with the corresponding Gaussians, allowing each primitive to jointly carry geometry- and semantics-related attributes. In this way, heterogeneous supervision originally defined in different representation spaces can be spatially aligned within a unified reconstructed 3D coordinate system, rather than being independently distilled in separate 2D feature spaces~\citep{kerbl2023gaussian,kerr2023lerf,zhou2024feature3dgs,qin2024langsplat}. In addition, Gaussian rendering naturally produces a coverage signal, which we use to restrict supervision to locations supported by valid geometry and non-negligible rendered Gaussian contributions.

Concretely, GaussianWAM constructs a Gaussian field from current multi-view observations and renders its geometric, semantic, and coverage signals back onto the token grid aligned with the WAM backbone. These dense signals are distilled into current-observation video latents
through training-time Gaussian distillation, encouraging WAM
representations to retain geometry-related and spatially aligned semantic
information. In other words, the enhanced representations better capture both \emph{where} objects are in 3D space and \emph{what} the corresponding visual regions represent. Crucially, the Gaussian field and all external teachers are used only during training and are completely removed at inference time. GaussianWAM therefore introduces no additional 3D reconstruction, semantic encoding, Gaussian rendering, or rollout overhead during deployment, preserving the original inference pipeline and computational cost of the underlying WAM.

Our contributions are:
\begin{itemize}[leftmargin=*,itemsep=1pt,topsep=2pt]
    \item We propose \textbf{GaussianWAM}, a general enhancement framework for WAMs that uses a Gaussian field as a unified 3D spatial carrier to organize complementary geometric and visual-semantic supervision for action-relevant latent representations.
    
    \item We introduce a Gaussian-field-based distillation mechanism that
    first binds geometric and visual-semantic knowledge within a unified 3D
    spatial representation, and then distills the rendered semantic, depth,
    and coverage signals into current-observation WAM representations.
    \item We adopt a \textbf{training-only enhancement} strategy that completely removes the Gaussian field and external teachers at inference, preserving the original WAM inference pipeline and computational cost. We extensively validate GaussianWAM on two representative WAM architectures, FastWAM and Cosmos Policy, and observe consistent improvements across \textbf{LIBERO, LIBERO-Plus, RoboTwin, and real-world robotic experiments}, demonstrating its effectiveness, robustness, and generalization ability.
\end{itemize}

\section{Related Work}
\label{sec:related-work}

\paragraph{World-action models and video-action models.}
World-Action Models (WAMs) and video-action models augment robot policies with predictive world modeling, jointly learning future visual dynamics and action generation. Representative methods such as Fast-WAM~\citep{yuan2026fastwam}, Cosmos Policy~\citep{kim2026cosmos}, LingBot~\citep{li2026lingbot}, and GigaWorld-Policy~\citep{giga2026gigaworld} demonstrate that future observation or latent dynamics modeling can provide effective representation-learning signals for robotic manipulation. Compared with purely reactive VLA policies, these approaches enable the policy to exploit temporal dynamics and implicit physical priors learned from video generation. However, their world representations are still predominantly modeled in RGB image space or 2D video latent space, leaving the resulting representations weakly grounded in explicit 3D structure and spatially aligned visual semantics. GaussianWAM retains the original WAM architecture and enhances its internal representations through structured training-time supervision without modifying the inference pipeline.

\paragraph{Geometry- and semantic-aware world-action learning.}
Recent studies have begun to incorporate structured geometric and semantic priors into WAMs to improve spatial understanding and action generation~\citep{li2026wam4d,zhang2026mecowam,ma2026geosem,yan2026svam,yang2026fourDwam,guo2026xwam}. These methods explore complementary directions including explicit 3D/4D representations, geometric foresight, spatial features, and auxiliary geometry or semantic prediction. In particular, GeoSem-WAM~\citep{ma2026geosem} highlights the complementary role of geometry and semantics in learning structured world representations beyond RGB prediction. Despite these advances, geometric and semantic knowledge is typically introduced through task-specific representations, prediction branches, or separate supervision objectives, without an explicit shared spatial carrier that associates both types of information with the same physical 3D structures. GaussianWAM instead unifies geometric and visual-semantic knowledge within a common 3D Gaussian field and distills the resulting structured representation into action-relevant WAM latents.

\paragraph{3D Gaussian fields for representation enhancement.}
3D Gaussian Splatting (3DGS)~\citep{kerbl2023gaussian} provides an explicit and efficient 3D scene representation in which Gaussian primitives encode spatial attributes such as position, scale, rotation, and opacity. Beyond appearance reconstruction, recent works have extended Gaussian primitives to carry high-dimensional features from pretrained foundation models. Feature 3DGS~\citep{zhou2024feature3dgs} distills 2D foundation-model features into Gaussian primitives, while LangSplat~\citep{qin2024langsplat} constructs language-aware Gaussian fields for open-vocabulary 3D understanding.  Gaussian world models such as GWM~\citep{lu2025gwm} and ManiGaussian~\citep{lu2024manigaussian} further explore Gaussian representations for robotic manipulation and dynamic scene modeling. More recently, Feature4X~\citep{zhou2025feature4x} demonstrates that heterogeneous features from visual and video foundation models can be lifted into a unified dynamic Gaussian representation. These works establish Gaussian primitives as a flexible spatial carrier that jointly preserves explicit geometry while accommodating rich semantic features. Building on this property, GaussianWAM uses a 3D Gaussian field to spatially bind geometric and visual-semantic knowledge and render dense supervision for WAM representation enhancement. Unlike prior Gaussian feature fields that primarily serve as scene representations for downstream 3D/4D perception and interaction, our Gaussian field is used only as a training-time knowledge carrier and is completely removed during policy inference.
\section{Method}
\label{sec:method}

\subsection{Overview}
\label{sec:method_overview}

\begin{figure*}[t]
    \centering
    \includegraphics[width=\textwidth]{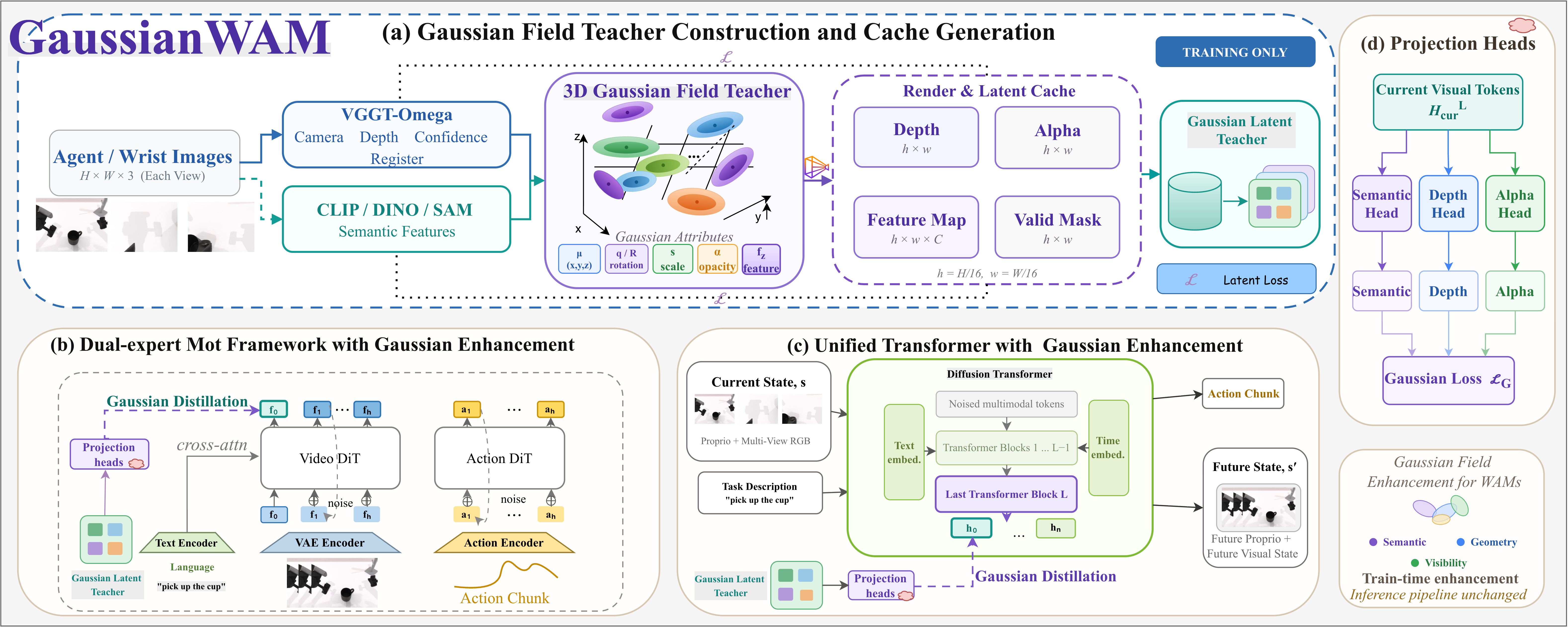}
\caption{
\textbf{Overview of GaussianWAM.}
(a) A 3D Gaussian teacher is constructed from synchronized multi-view
observations and rendered into semantic, depth, and alpha targets,
together with a validity mask, which are cached for training.
(b) In FastWAM-style dual-expert MoT models, Gaussian supervision is applied
to current-observation video representations.
(c) In Cosmos-Policy-style unified Transformers, the same supervision is
applied to current-observation visual tokens in the shared backbone.
(d) Lightweight semantic, depth, and alpha heads distill the cached Gaussian targets into WAM visual representations.
All Gaussian-related modules are removed at inference, preserving the
original WAM forward path and computational cost.
}
    \label{fig:framework}
\end{figure*}

We propose \textbf{GaussianWAM}, a general training-time 3D Gaussian
enhancement framework for World-Action Models.
As shown in Fig.~\ref{fig:framework}, GaussianWAM first constructs
an offline 3D Gaussian teacher from current multi-view observations.
The Gaussian field provides a unified spatial representation that
combines 3D geometry, visual-semantic features, and rendering-derived
coverage information.
These signals are rendered onto spatial grids aligned with the visual
representations of the WAM and distilled into action-relevant visual
representations during policy training.

The Gaussian teacher is used only during training.
At inference time, all foundation-model teachers, Gaussian construction
and rendering modules, and auxiliary prediction heads are removed.
The enhanced policy therefore follows exactly the same forward path
as the original WAM.
We instantiate GaussianWAM on both a FastWAM-style dual-expert
MoT architecture~\citep{yuan2026fastwam} and a Cosmos-Policy-style
unified DiT~\citep{kim2026cosmos}.

\subsection{Gaussian Teacher Construction}
\label{sec:gaussian_teacher}

For each training sample, we construct a dense 3D Gaussian teacher
from synchronized multi-view observations
$\{I^v\}_{v=1}^{V}$.
The teacher takes only visual observations as input, while language
instructions and proprioceptive states remain inputs to the policy.

\paragraph{Geometry extraction.}
We employ a frozen geometry foundation model,
VGGT-Omega~\citep{wang2025vggt}, to estimate dense depth,
depth confidence, and camera parameters:
\begin{equation}
    \{D^v,C^v,K^v,E^v\}_{v=1}^{V}
    =
    \mathrm{VGGT}(\{I^v\}_{v=1}^{V}),
\end{equation}
where $D^v$ and $C^v$ denote the depth and depth-confidence maps,
while $K^v$ and $E^v$ denote the corresponding camera intrinsics
and extrinsics.

The estimated depth and confidence maps are resized to a common
$14\times14$ teacher grid for each view, with the camera intrinsics
scaled accordingly.
Invalid or low-confidence estimates are filtered to obtain a
geometry-valid mask $M_{\mathrm{geo}}^v$.
The remaining valid pixels are then back-projected into 3D space
using the corresponding depth and camera parameters.

\paragraph{Visual-semantic feature extraction.}
We additionally extract dense visual-semantic features using a frozen
CLIP image encoder~\citep{radford2021clip}.
Specifically, we use CLIP ViT-B/16 in all experiments.
The patch tokens are reshaped into a spatial feature map and resized
to the same teacher grid.
The resulting CLIP features are projected to a 64-dimensional
representation, yielding $F_{\mathrm{sem}}^v$ for each view.
By placing geometric cues and visual-semantic features on the same
spatial grid, we associate the two types of information at
corresponding image locations before lifting them into the Gaussian
field.
We use CLIP in our implementation, while the same construction is
compatible in principle with other dense visual foundation encoders,
such as DINO~\citep{oquab2023dinov2} or
SAM~\citep{kirillov2023sam}.

\paragraph{Gaussian field initialization.}
Using the geometry-valid 3D points, we initialize a Gaussian field
$\mathcal{G}=\{g_i\}_{i=1}^{N}$.
We adopt dense initialization with spatial stride $1$ on the teacher
grid, such that every geometry-valid location initializes one
Gaussian primitive:
\begin{equation}
    g_i =
    \{\mathbf{x}_i,\mathbf{r}_i,\mathbf{s}_i,o_i,\mathbf{z}_i\},
\end{equation}
where $\mathbf{x}_i\in\mathbb{R}^{3}$ denotes the 3D center,
$\mathbf{r}_i$ represents the Gaussian rotation,
$\mathbf{s}_i\in\mathbb{R}^{3}$ denotes the spatial scale,
$o_i$ denotes opacity,
and $\mathbf{z}_i\in\mathbb{R}^{64}$ stores the visual-semantic
feature initialized from the corresponding location in
$F_{\mathrm{sem}}^v$.
This construction binds explicit 3D geometry and visual-semantic
information to the same Gaussian primitives within a common
3D coordinate system.

\paragraph{Multi-view Gaussian fitting.}
We render the Gaussian field into each observed camera view using a
differentiable depth-aware Gaussian renderer $\mathcal{R}$:
\begin{equation}
    (\hat{F}^v,\hat{D}^v,\hat{A}^v)
    =
    \mathcal{R}(\mathcal{G};K^v,E^v),
\end{equation}
where $\hat{F}^v$, $\hat{D}^v$, and $\hat{A}^v$ denote the rendered
visual-semantic feature map, depth map, and accumulated Gaussian
coverage map, respectively.
Specifically, $\mathcal{R}$ transforms the Gaussian primitives into
the target camera frame, projects them onto the teacher grid using
$K^v$, and performs depth-aware, opacity-weighted soft splatting.
The projected Gaussian contributions are aggregated to obtain the
rendered feature and depth maps, while their accumulated weights
define the coverage map $\hat{A}^v$.

We optimize the Gaussian field jointly across all observed views using
\begin{equation}
    \mathcal{L}_{\mathrm{fit}}
    =
    \lambda_s\mathcal{L}^{\mathrm{fit}}_{\mathrm{sem}}
    +
    \lambda_d\mathcal{L}^{\mathrm{fit}}_{\mathrm{depth}}
    +
    \lambda_c\mathcal{L}_{\mathrm{cov}}
    +
    \mathcal{L}_{\mathrm{reg}},
\end{equation}
where $\mathcal{L}^{\mathrm{fit}}_{\mathrm{sem}}$ aligns
$\hat{F}^v$ with $F_{\mathrm{sem}}^v$ using cosine distance, and
$\mathcal{L}^{\mathrm{fit}}_{\mathrm{depth}}$ aligns
$\hat{D}^v$ with $D^v$ using an $\ell_1$ loss.
The coverage objective $\mathcal{L}_{\mathrm{cov}}$ encourages high
Gaussian coverage at geometry-valid locations while suppressing
spurious coverage outside valid regions.
$\mathcal{L}_{\mathrm{reg}}$ denotes lightweight regularization on
the Gaussian parameters, including scale regularization and penalties
on excessive center drift and, when optimized, deviation of semantic
features from their initialization.
The fitting stage optimizes the selected Gaussian parameters,
including their centers, rotations, scales, and opacities, with semantic features
optionally refined during fitting.
Each per-sample Gaussian field is optimized for
$N_{\mathrm{fit}}$ iterations, where we set
$N_{\mathrm{fit}}=50$.

\paragraph{Offline teacher cache.}
After fitting, we render the optimized Gaussian field and store
\begin{equation}
    \mathcal{T}
    =
    \{
    T_{\mathrm{sem}},
    T_{\mathrm{depth}},
    T_{\alpha},
    T_{\mathrm{valid}}
    \}.
\end{equation}
Here, $T_{\mathrm{sem}}$ is the visual-semantic feature map rendered
from the optimized Gaussian field,
$T_{\mathrm{depth}}$ is the rendered depth,
and $T_{\alpha}$ represents accumulated Gaussian coverage.
We define the final valid mask as
\begin{equation}
    T_{\mathrm{valid}}
    =
    M_{\mathrm{geo}}
    \cap
    M_{\mathrm{render}},
    \qquad
    M_{\mathrm{render}}
    =
    \mathbb{I}[T_{\alpha}>\tau_{\alpha}],
\end{equation}
where we set $\tau_{\alpha}=10^{-4}$ in all experiments.
This small threshold filters locations with negligible accumulated
Gaussian coverage, such that downstream supervision is applied only
at locations supported by both reliable geometry and valid Gaussian
rendering.

For WAMs that jointly process multiple camera views, the per-view
rendered targets are further resized and composed according to the
same multi-view spatial layout used by the policy representation.
The resulting semantic, depth, alpha, and validity maps are cached
offline and directly reused during policy training, avoiding repeated
foundation-model inference and Gaussian fitting.
\subsection{Gaussian Distillation}
\label{sec:gaussian_distillation}

GaussianWAM distills the cached Gaussian representation into the
internal visual representation of a WAM.
Let $H_{\mathrm{cur}}^{L}$ denote the final-layer hidden tokens
corresponding to the current visual observations.
Their exact source depends on the underlying WAM architecture and is
described in Secs.~\ref{sec:fastwam_instantiation}
and~\ref{sec:cosmos_instantiation}.

We attach three lightweight auxiliary prediction heads,
collectively denoted as
\begin{equation}
    \Phi_G =
    \{
    \phi_{\mathrm{sem}},
    \phi_{\mathrm{depth}},
    \phi_{\alpha}
    \},
\end{equation}
to predict the cached Gaussian targets:
\begin{equation}
\begin{aligned}
    \hat{T}_{\mathrm{sem}}
        &= \phi_{\mathrm{sem}}(H_{\mathrm{cur}}^{L}), \\
    \hat{T}_{\mathrm{depth}}
        &= \phi_{\mathrm{depth}}(H_{\mathrm{cur}}^{L}), \\
    \hat{T}_{\alpha}
        &= \sigma\!\left(\phi_{\alpha}(H_{\mathrm{cur}}^{L})\right),
\end{aligned}
\end{equation}

The Gaussian distillation objective is
\begin{equation}
    \mathcal{L}_{G}
    =
    \lambda_{\mathrm{sem}}\mathcal{L}_{\mathrm{sem}}
    +
    \lambda_{\mathrm{depth}}\mathcal{L}_{\mathrm{depth}}
    +
    \lambda_{\alpha}\mathcal{L}_{\alpha},
\end{equation}
where all losses are evaluated only over $T_{\mathrm{valid}}$.
Let $\Omega_{\mathrm{valid}}=\{p\mid T_{\mathrm{valid}}(p)=1\}$
denote the valid locations on the WAM-aligned token grid. We use
masked cosine distance for visual-semantic alignment:
\begin{equation}
\mathcal{L}_{\mathrm{sem}}
=
\frac{1}{|\Omega_{\mathrm{valid}}|}
\sum_{p\in\Omega_{\mathrm{valid}}}
\left(
1-
\frac{
\hat{T}_{\mathrm{sem}}(p)^\top T_{\mathrm{sem}}(p)
}{
\|\hat{T}_{\mathrm{sem}}(p)\|_2
\|T_{\mathrm{sem}}(p)\|_2+\epsilon
}
\right),
\end{equation}
where $\epsilon$ is a small constant for numerical stability.
and masked $\ell_1$ losses for depth and Gaussian coverage:
\begin{equation}
\mathcal{L}_{\mathrm{depth}}
=
\frac{1}{|\Omega_{\mathrm{valid}}|}
\sum_{p\in\Omega_{\mathrm{valid}}}
\left|
\hat{T}_{\mathrm{depth}}(p)-T_{\mathrm{depth}}(p)
\right|,
\end{equation}
\begin{equation}
\mathcal{L}_{\alpha}
=
\frac{1}{|\Omega_{\mathrm{valid}}|}
\sum_{p\in\Omega_{\mathrm{valid}}}
\left|
\hat{T}_{\alpha}(p)-T_{\alpha}(p)
\right|.
\end{equation}

The overall training objective is
\begin{equation}
    \mathcal{L}_{\mathrm{train}}
    =
    \mathcal{L}_{\mathrm{WAM}}
    +
    \lambda_G\mathcal{L}_{G},
\end{equation}
where $\mathcal{L}_{\mathrm{WAM}}$ denotes the original training
objective of the underlying WAM.

The Gaussian teacher and auxiliary prediction heads are used only
during training.
At inference time, they are completely removed, leaving the original
WAM forward path and inference cost unchanged.

\subsection{FastWAM-Style Dual-Expert MoT}
\label{sec:fastwam_instantiation}

FastWAM-style models~\citep{yuan2026fastwam} employ separate video
and action experts coupled through a Mixture-of-Transformers (MoT)
architecture.
The video expert maintains the visual world representation, which is
accessed by the action expert through MoT interaction.
We therefore apply Gaussian distillation to the final-layer video
representation.

Let $H_{\mathrm{vid}}^{L}$ denote the final-layer video hidden states.
We select the spatial tokens corresponding to the current visual
observations:
\begin{equation}
    H_{\mathrm{cur}}^{L}
    =
    \mathrm{Select}_{\mathrm{cur}}
    (H_{\mathrm{vid}}^{L}),
\end{equation}
restore their spatial organization according to the visual token grid,
and apply the Gaussian distillation objective defined in
Sec.~\ref{sec:gaussian_distillation}.

Through the MoT interaction, the geometry- and semantic-enhanced video
representation is directly available to the action expert.
Gaussian supervision can therefore influence action learning without
introducing an additional action pathway or modifying the original
FastWAM inference procedure.

\subsection{Cosmos-Policy-Style Unified DiT}
\label{sec:cosmos_instantiation}

Cosmos-Policy-style models~\citep{kim2026cosmos} instead process
visual observations, world-related variables, and action-related
tokens within a unified diffusion Transformer.
Since there is no separate video expert, Gaussian distillation is
applied directly to the final-layer hidden tokens corresponding to
the current visual observations.

Let $H^{L}$ denote the final Transformer hidden states.
We extract the current-observation tokens as
\begin{equation}
    H_{\mathrm{cur}}^{L}
    =
    \mathrm{Select}_{\mathrm{cur}}(H^{L}),
\end{equation}
restore their spatial organization, and apply the same Gaussian
distillation mechanism from Sec.~\ref{sec:gaussian_distillation}.

Because these visual tokens belong to the shared Transformer backbone,
Gaussian supervision directly shapes the latent representation shared
by world and action modeling.
As with FastWAM, all Gaussian-related modules are removed after
training, preserving the original Cosmos Policy inference path.
\section{Experiments}
\label{sec:experiments}

\subsection{Experimental Setup}
\label{sec:exp_setup}

\begin{table*}[t]
\centering
\small
\caption{
Main results on LIBERO and LIBERO-Plus.
We report success rate (\%).
All models are trained on LIBERO and evaluated on both LIBERO and
LIBERO-Plus, where LIBERO-Plus serves as a zero-shot evaluation under
distribution shifts.
Emb. PT. indicates large-scale embodied pretraining.
Our FastWAM baseline and its GaussianWAM variant are trained for
70k optimization steps, while the Cosmos Policy baseline and its
GaussianWAM variant are trained for 5k iterations.
}
\label{tab:main_results}

\resizebox{\textwidth}{!}{%
\begin{tabular}{llcccccccccc}
\toprule
Method & Type & Emb. PT.
& LIBERO
& Camera & Robot & Lang. & Light & BG & Noise & Layout
& Overall \\
\midrule

\multicolumn{12}{l}{\textit{Vision-Language-Action Models}} \\
\midrule

UniVLA~\citep{bu2025univla}
& VLA & Yes
& 95.20
& 1.80 & 46.20 & 69.60 & 69.00 & 90.70 & 21.20 & 31.90
& 43.90 \\

$\pi_0$~\citep{black2024pi0}
& VLA & Yes
& 94.20
& 13.80 & 6.00 & 58.80 & 85.00 & 90.70 & 79.00 & 68.90
& 54.60 \\

$\pi_0$-FAST~\citep{pertsch2025fast}
& VLA & Yes
& 85.50
& 65.10 & 21.60 & 61.00 & 73.20 & 97.70 & 74.40 & 68.80
& 64.20 \\

OpenVLA-OFT~\citep{kim2025openvlaoft}
& VLA & Yes
& 97.10
& 56.40 & 31.90 & 79.50 & 88.70 & \textbf{97.30} & 75.80 & 74.20
& 70.00 \\

X-VLA~\citep{zheng2025xvla}
& VLA & Yes
& 98.10
& 23.40 & \textbf{89.70} & 75.70 & 88.20 & 96.00 & 62.70 & 71.80
& 70.46 \\

Spatial Forcing~\citep{li2025spatialforcing}
& VLA & Yes
& 98.50
& 20.10 & 13.40 & 40.90 & 29.10 & 33.40 & 25.70 & 39.30
& 28.52 \\

\midrule
\multicolumn{12}{l}{\textit{Geometry-, Semantic-, and Structured World-Action Models}} \\
\midrule

GeoSem-WAM~\citep{ma2026geosem}
& WAM & No
& 98.55
& -- & -- & -- & -- & -- & -- & --
& -- \\

SG-WAM~\citep{zhao2026sgwam}
& WAM & No
& 98.50
& 58.60 & 48.90 & 81.40 & 89.80 & 86.10 & 80.70 & 74.20
& 73.00 \\

DreamWAM~\citep{yuan2026dreamwam}
& WAM & No
& \textbf{98.90}
& 53.78 & 63.61 & \textbf{94.80} & \textbf{96.67}
& 71.56 & 67.15 & \textbf{80.72}
& 74.61 \\

ST-WAM~\citep{wang2026stwam}
& WAM & No
& 98.70
& 55.40 & 60.10 & 79.30 & 93.00
& 74.20 & 79.50 & 74.30
& 72.80 \\

4D-WAM~\citep{yang2026fourDwam}
& WAM & No
& 98.60
& 45.15 & 64.26 & 90.63 & 94.29
& 57.71 & 69.08 & 79.21
& 71.01 \\

\midrule
\multicolumn{12}{l}{\textit{GaussianWAM on FastWAM}} \\
\midrule

FastWAM~\citep{yuan2026fastwam}
& WAM & No
& 96.60
& 25.63 & 42.88 & 73.88 & 61.70
& 54.88 & 43.55 & 66.80
& 52.05 \\

FastWAM + GaussianWAM
& WAM & No
& 97.60 {\color{green!50!black}{(+1.00)}}
& 54.66 {\color{green!50!black}{(+29.03)}}
& 61.61 {\color{green!50!black}{(+18.73)}}
& 75.41 {\color{green!50!black}{(+1.53)}}
& 90.81 {\color{green!50!black}{(+29.11)}}
& 58.18 {\color{green!50!black}{(+3.30)}}
& 83.01 {\color{green!50!black}{(+39.46)}}
& 76.72 {\color{green!50!black}{(+9.92)}}
& 71.29 {\color{green!50!black}{(+19.24)}} \\

\midrule
\multicolumn{12}{l}{\textit{GaussianWAM on Cosmos Policy}} \\
\midrule

Cosmos Policy~\citep{kim2026cosmos}
& WAM & No
& 98.50
& 69.67 & 42.39 & 86.26 & 90.89
& 77.51 & 75.58 & 65.23
& 71.52 \\

Cosmos Policy + GaussianWAM
& WAM & No
& 98.60 {\color{green!50!black}{(+0.10)}}
& \textbf{79.11} {\color{green!50!black}{(+9.44)}}
& 56.52 {\color{green!50!black}{(+14.13)}}
& 92.18 {\color{green!50!black}{(+5.92)}}
& 89.84 {\color{red}{(-1.05)}}
& 66.17 {\color{red}{(-11.34)}}
& \textbf{86.26} {\color{green!50!black}{(+10.68)}}
& 70.55 {\color{green!50!black}{(+5.32)}}
& \textbf{77.30} {\color{green!50!black}{(+5.78)}} \\

\bottomrule
\end{tabular}%
}
\end{table*}

\paragraph{Benchmarks.}
We evaluate GaussianWAM on LIBERO~\citep{liu2024libero},
LIBERO-Plus~\citep{fei2025liberoplus}, RoboTwin, and real-world
robotic manipulation tasks.
LIBERO evaluates standard language-conditioned manipulation, while
LIBERO-Plus introduces diverse distribution shifts in camera viewpoint,
robot appearance, language instruction, illumination, background,
visual noise, and scene layout.
RoboTwin is a bimanual manipulation benchmark evaluated under both
\emph{Clean} and \emph{Random} settings, where the latter introduces
stronger scene and visual randomization to test robustness under
distribution shifts.
Our real-world experiments further evaluate whether the learned
representation improvements transfer beyond simulation.
We report task success rate (\%) as the main evaluation metric.

\paragraph{Backbones and training.}
We instantiate GaussianWAM on two representative WAM architectures:
FastWAM~\citep{yuan2026fastwam}, based on a dual-expert
Mixture-of-Transformers (MoT) architecture, and
Cosmos Policy~\citep{kim2026cosmos}, based on a unified diffusion
Transformer.
For both architectures, Gaussian distillation is applied to the
final-layer visual representation corresponding to the current
observations.

For FastWAM, we train both the baseline and its GaussianWAM variant
for 70k optimization steps on 8 NVIDIA A100 GPUs, with a per-GPU
batch size of 2 and gradient accumulation over 2 steps, resulting in
an effective batch size of 32.
We use a learning rate of $1\times10^{-4}$ with cosine decay and
a weight decay of $1\times10^{-2}$.
The Gaussian distillation losses are weighted by
$\lambda_{\mathrm{sem}}=0.01$,
$\lambda_{\mathrm{depth}}=0.01$, and
$\lambda_{\alpha}=0.005$.

For Cosmos Policy, we use 8 GPUs with a local batch size of 30 and
gradient accumulation over 8 steps, resulting in an effective batch
size of 1920, following its LIBERO training configuration.
Both the baseline and its GaussianWAM variant are trained for 5k
iterations.
For each backbone, the baseline and GaussianWAM variant use the same
training data, backbone configuration, and optimization budget, with
Gaussian distillation introduced only during training.

\subsection{Main Results}
\label{sec:main_results}

\paragraph{LIBERO and LIBERO-Plus.}
Table~\ref{tab:main_results} reports the main results on LIBERO and
LIBERO-Plus.
GaussianWAM consistently improves the underlying WAM backbones.
For FastWAM, standard LIBERO success increases from 96.6\% to 97.6\%,
while the official LIBERO-Plus overall score improves substantially
from 52.05\% to 71.29\%.
Particularly large gains are observed under camera
($+29.03$), lighting ($+29.11$), noise ($+39.46$), and robot
($+18.73$) shifts, suggesting that structured Gaussian supervision
provides substantially stronger spatial grounding under challenging
visual variations.

The improvement is also consistent on a distinct WAM architecture.
Under the same 5k-iteration training budget, GaussianWAM improves
Cosmos Policy from 71.52\% to 77.30\% overall.
Notable gains are obtained for camera, robot, language, noise, and
layout shifts.
These results indicate that the proposed enhancement is not specific
to the dual-expert FastWAM architecture, but can also benefit a unified world-action Transformer.

\paragraph{RoboTwin and real-world evaluation.}

\begin{wraptable}{l}{0.48\textwidth}
    \centering
    \small
    \caption{
    Results on the 15-task RoboTwin 2.0 Clean-to-Random benchmark.
    Clean and Random denote the clean and randomized evaluation settings,
    respectively. We report average success rate (\%).
    }
    \label{tab:robotwin_results}
    \resizebox{\linewidth}{!}{%
    \begin{tabular}{lcc}
    \toprule
    Method & Clean & Random \\
    \midrule
    DP & 34.07 & 0.33 \\
    ACT & 34.20 & 4.00 \\
    DP3 & 59.87 & 3.80 \\
    \midrule
    FastWAM & 70.50 & 0.70 \\
    FastWAM + GaussianWAM & \textbf{70.50} & 1.60 \\
    \midrule
    Cosmos Policy & 34.40 & 7.10 \\
    Cosmos Policy + GaussianWAM & 68.90 & \textbf{14.40} \\
    \bottomrule
    \end{tabular}%
    }
\end{wraptable}

We further evaluate GaussianWAM on RoboTwin and real-world robotic
manipulation tasks.
These experiments complement LIBERO by introducing different scene
distributions, object configurations, and embodiment conditions.
As shown in Table~\ref{tab:robotwin_results}, GaussianWAM matches or
improves the corresponding base WAM on RoboTwin, with particularly
clear gains under the Random setting.
For FastWAM, GaussianWAM preserves the Clean performance at 70.50\%
while improving Random success from 0.70\% to 1.60\%.
For Cosmos Policy, GaussianWAM improves Clean success from 34.40\%
to 68.90\% and Random success from 7.10\% to 14.40\%.
These results demonstrate that the learned representation enhancement
transfers beyond the LIBERO environment and provides stronger
robustness under randomized evaluation conditions.

\paragraph{Real-world manipulation.}
We further evaluate GaussianWAM on a bimanual real-robot platform
consisting of two UR7e robotic arms.
We consider two manipulation tasks: lifting a vase to contact the edge
of a red plate, and placing a white cup at the designated center before
inserting a purple cylinder into it.
For each task, we collect 100 demonstrations and evaluate each method
over 20 independent trials under every evaluation setting.
A trial is considered successful only when all task-specific objectives
are completed.

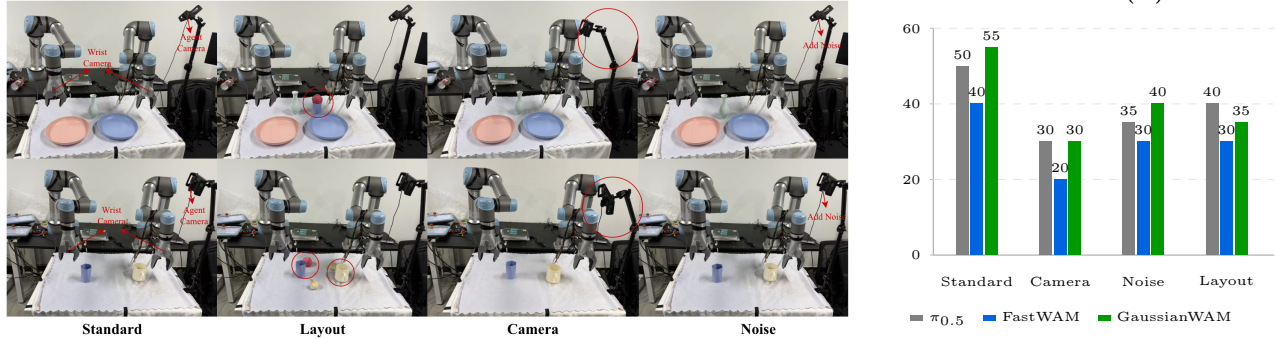
\begin{figure*}[t]
    \centering

    \begin{minipage}[c][5cm][c]{0.70\textwidth}
        \centering
        \includegraphics[
            width=\linewidth,
            height=5cm,
            keepaspectratio
        ]{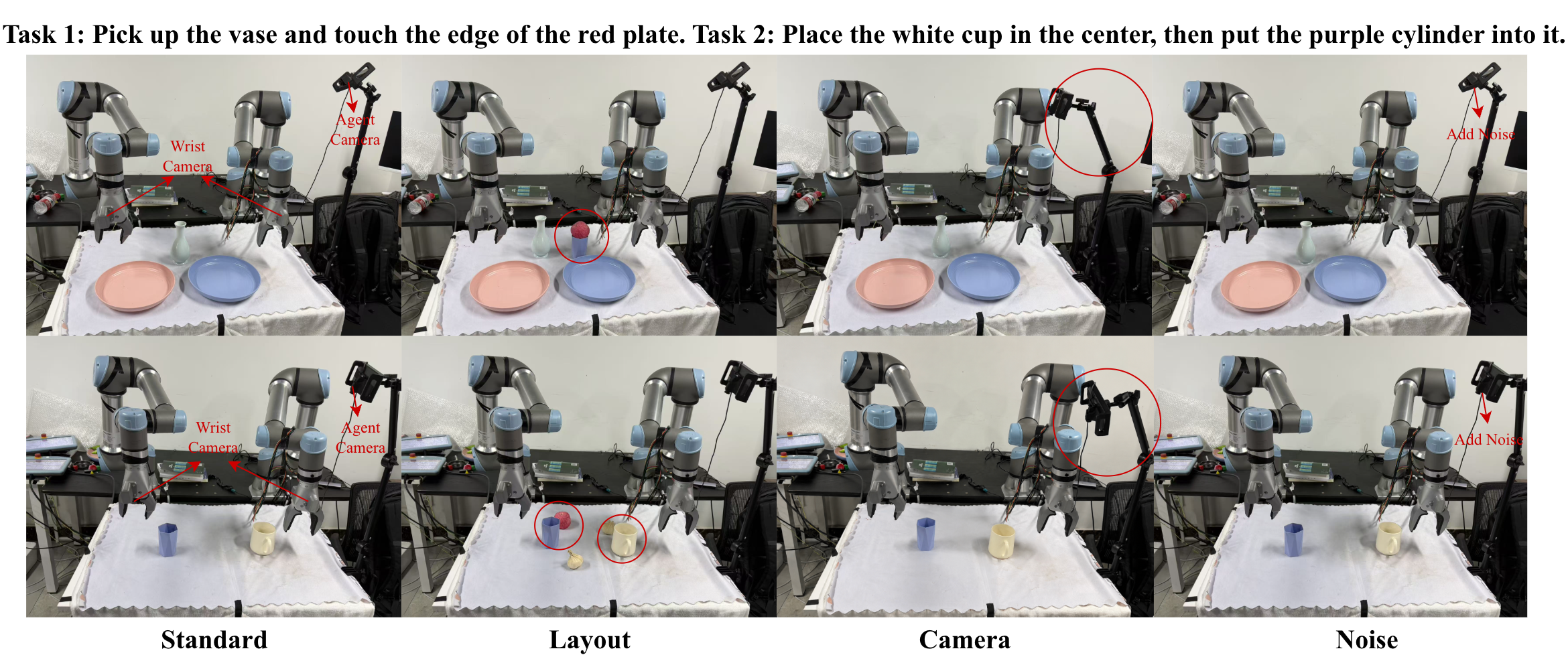}
    \end{minipage}
    \hfill
    \begin{minipage}[c][5cm][c]{0.28\textwidth}
        \centering

        \scriptsize
        \textbf{Success Rate (\%)}\par
        \vspace{4pt}

        \begin{tikzpicture}[x=0.55cm,y=0.050cm]

            \draw[gray!70] (0,0) -- (0,60);
            \draw[gray!70] (0,0) -- (8.2,0);

            \foreach \y in {0,20,40,60}{
                \draw[gray!50] (-0.08,\y) -- (0,\y);
                \node[font=\tiny,anchor=east]
                    at (-0.12,\y) {\y};
            }

            \foreach \y in {20,40,60}{
                \draw[gray!20,dashed]
                    (0,\y) -- (8.2,\y);
            }

            \fill[gray]
                (0.55,0) rectangle (0.85,50);
            \fill[metablue]
                (0.90,0) rectangle (1.20,40);
            \fill[green!60!black]
                (1.25,0) rectangle (1.55,55);

            \node[font=\tiny,above] at (0.70,50) {50};
            \node[font=\tiny,above] at (1.05,40) {40};
            \node[font=\tiny,above] at (1.40,55) {55};

            \node[font=\tiny] at (1.05,-7) {Standard};

            \fill[gray]
                (2.55,0) rectangle (2.85,30);
            \fill[metablue]
                (2.90,0) rectangle (3.20,20);
            \fill[green!60!black]
                (3.25,0) rectangle (3.55,30);

            \node[font=\tiny,above] at (2.70,30) {30};
            \node[font=\tiny,above] at (3.05,20) {20};
            \node[font=\tiny,above] at (3.40,30) {30};

            \node[font=\tiny] at (3.05,-7) {Camera};

            \fill[gray]
                (4.55,0) rectangle (4.85,35);
            \fill[metablue]
                (4.90,0) rectangle (5.20,30);
            \fill[green!60!black]
                (5.25,0) rectangle (5.55,40);

            \node[font=\tiny,above] at (4.70,35) {35};
            \node[font=\tiny,above] at (5.05,30) {30};
            \node[font=\tiny,above] at (5.40,40) {40};

            \node[font=\tiny] at (5.05,-7) {Noise};

            \fill[gray]
                (6.55,0) rectangle (6.85,40);
            \fill[metablue]
                (6.90,0) rectangle (7.20,30);
            \fill[green!60!black]
                (7.25,0) rectangle (7.55,35);

            \node[font=\tiny,above] at (6.70,40) {40};
            \node[font=\tiny,above] at (7.05,30) {30};
            \node[font=\tiny,above] at (7.40,35) {35};

            \node[font=\tiny] at (7.05,-7) {Layout};

        \end{tikzpicture}

        \vspace{4pt}

        \tiny
        \textcolor{gray}{\rule{0.16cm}{2.2pt}}
        $\pi_{0.5}$\quad
        \textcolor{metablue}{\rule{0.16cm}{2.2pt}}
        FastWAM\quad
        \textcolor{green!60!black}{\rule{0.16cm}{2.2pt}}
        GaussianWAM

    \end{minipage}

    \caption{
    Real-world evaluation and success rates.
    The left panel shows two bimanual tasks under standard, layout,
    camera, and visual-noise variations.
    The right panel shows success rates for $\pi_{0.5}$, FastWAM,
    and GaussianWAM under the four settings.
    Each task uses 100 demonstrations and 20 evaluation trials per setting.
    }
    \label{fig:real_robot}
\end{figure*}

Figure~\ref{fig:real_robot} shows the evaluation setup under four
conditions: standard, layout, camera, and visual-noise variations.
GaussianWAM improves the average success rate of FastWAM from
30.00\% to 40.00\%.
These results demonstrate that the representation enhancement learned
through training-time Gaussian distillation remains effective under
real-world distribution shifts, while preserving the original WAM
inference pipeline at deployment.

\subsection{Ablation Studies}
\label{sec:ablation_study}

\paragraph{Effect of Gaussian-field unification.}
We first compare GaussianWAM with a direct 2D distillation baseline,
where CLIP semantic features and VGGT depth are independently distilled
into the WAM representation without constructing a Gaussian field.
Direct CLIP + VGGT distillation already provides a strong improvement
over FastWAM, increasing the official LIBERO-Plus overall success rate
from 52.05\% to 69.37\%.
Notably, GaussianWAM without alpha supervision further reaches 70.99\%,
outperforming direct CLIP + VGGT distillation by 1.62 points.
This result supports the benefit of organizing heterogeneous teacher
signals through a shared 3D Gaussian representation, beyond direct
multi-teacher supervision.
Adding rendering-derived alpha supervision further improves the full
model to 71.29\%, providing an additional signal of reliable 3D
spatial support.

\paragraph{Effect of Gaussian supervision components.}
We next study the contribution of semantic, depth, and alpha supervision
on LIBERO-Plus.
As shown in Table~\ref{tab:task_ablation}, removing any individual
component degrades the overall performance.
Removing semantic or alpha supervision reduces the official overall
success rate from 71.29\% to 68.06\% and 70.99\%, respectively,
highlighting the importance of both visual-semantic grounding and
rendering-derived 3D spatial support.
Removing depth supervision results in the largest performance drop,
reducing the overall success rate to 66.28\%, demonstrating the
importance of explicit geometric information for robust manipulation.
Together, these results show that geometry, visual semantics, and
Gaussian coverage provide complementary supervision for GaussianWAM.

\paragraph{Effect of supervision depth.}
We further investigate where Gaussian distillation should be applied
within the WAM backbone.
As shown in Table~\ref{tab:layer_ablation}, applying Gaussian
distillation to the final Transformer layer (layer 30) achieves the
best official overall performance of 71.29\%, compared with
67.86\% and 68.44\% at layers 10 and 20, respectively.
Although intermediate layers perform favorably under several individual
distribution shifts, final-layer supervision provides the strongest
overall performance.
We therefore apply Gaussian distillation to the final-layer visual
representation in our main configuration.

\begin{table*}[t]
\centering

\begin{minipage}[t]{0.48\textwidth}
\centering
\small
\captionof{table}{
Ablation of Gaussian-field unification and supervision components on
LIBERO-Plus with FastWAM.
Overall is computed over the official 10,030 evaluation tasks.
}
\label{tab:task_ablation}
\resizebox{\linewidth}{!}{%
\begin{tabular}{lcccccccc}
\toprule
Variant & Camera & Robot & Lang. & Light & BG & Noise & Layout & Overall \\
\midrule

FastWAM
& 25.63 & 42.88 & 73.88 & 61.70
& 54.88 & 43.55 & 66.80 & 52.05 \\

Direct CLIP + VGGT
& 48.41 & \textbf{63.74} & 69.75 & 88.53
& \textbf{61.71} & 82.14 & 74.36 & 69.37 \\

\midrule

Full
& 54.66 & 61.61 & \textbf{75.41} & \textbf{90.81}
& 58.18 & \textbf{83.01} & \textbf{76.72} & \textbf{71.29} \\

w/o Semantic
& 49.30 & 62.80 & 73.30 & 84.80
& 55.40 & 77.20 & 74.60 & 68.06 \\

w/o Depth
& 50.78 & 58.84 & 68.58 & 83.54
& 52.23 & 78.33 & 72.13 & 66.28 \\

w/o Alpha
& \textbf{58.85} & \textbf{63.74} & 72.61 & 87.74
& 59.29 & 79.51 & 76.20 & 70.99 \\

\bottomrule
\end{tabular}%
}
\end{minipage}
\hfill
\begin{minipage}[t]{0.48\textwidth}
\centering
\small
\captionof{table}{
Ablation of supervision depth on LIBERO-Plus with
FastWAM + GaussianWAM.
Layer 30 corresponds to the final Transformer layer.
Overall is computed over the official 10,030 evaluation tasks.
}
\label{tab:layer_ablation}
\resizebox{\linewidth}{!}{%
\begin{tabular}{lcccccccc}
\toprule
Layer & Camera & Robot & Lang. & Light & BG & Noise & Layout & Overall \\
\midrule

10
& 52.28 & \textbf{64.97} & 62.65 & 88.79
& 61.52 & 78.08 & 70.49 & 67.86 \\

20
& 51.91 & 62.77 & 69.81 & 86.95
& \textbf{64.50} & 75.20 & 72.00 & 68.44 \\

30
& \textbf{54.66} & 61.61 & \textbf{75.41} & \textbf{90.81}
& 58.18 & \textbf{83.01} & \textbf{76.72} & \textbf{71.29} \\

\bottomrule
\end{tabular}%
}
\end{minipage}

\end{table*}

\subsection{Qualitative Analysis}
\label{sec:qualitative}

We further analyze the geometric and visual-semantic representations
learned with GaussianWAM.

\begin{figure*}[t]
    \centering

    \begin{minipage}[t]{0.48\textwidth}
        \centering
        \includegraphics[width=\linewidth]
        {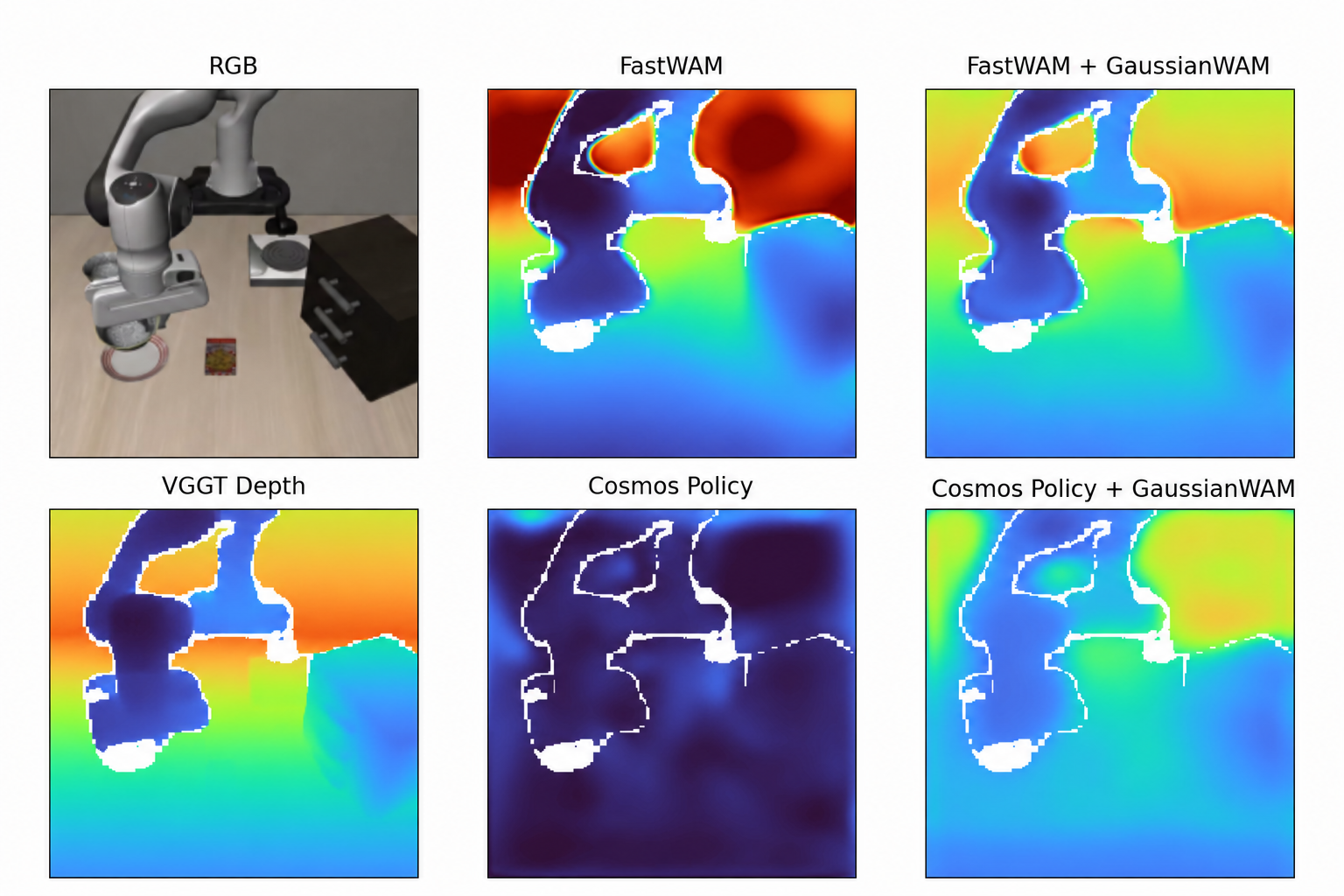}
        \captionof{figure}{
        Frozen-backbone depth probing of WAM visual representations.
        The same lightweight depth probe is applied to the base and
        GaussianWAM models using VGGT-Omega pseudo-depth as supervision.
        }
        \label{fig:depth_probe}
    \end{minipage}
    \hfill
    \begin{minipage}[t]{0.48\textwidth}
        \centering
        \includegraphics[width=\linewidth]
        {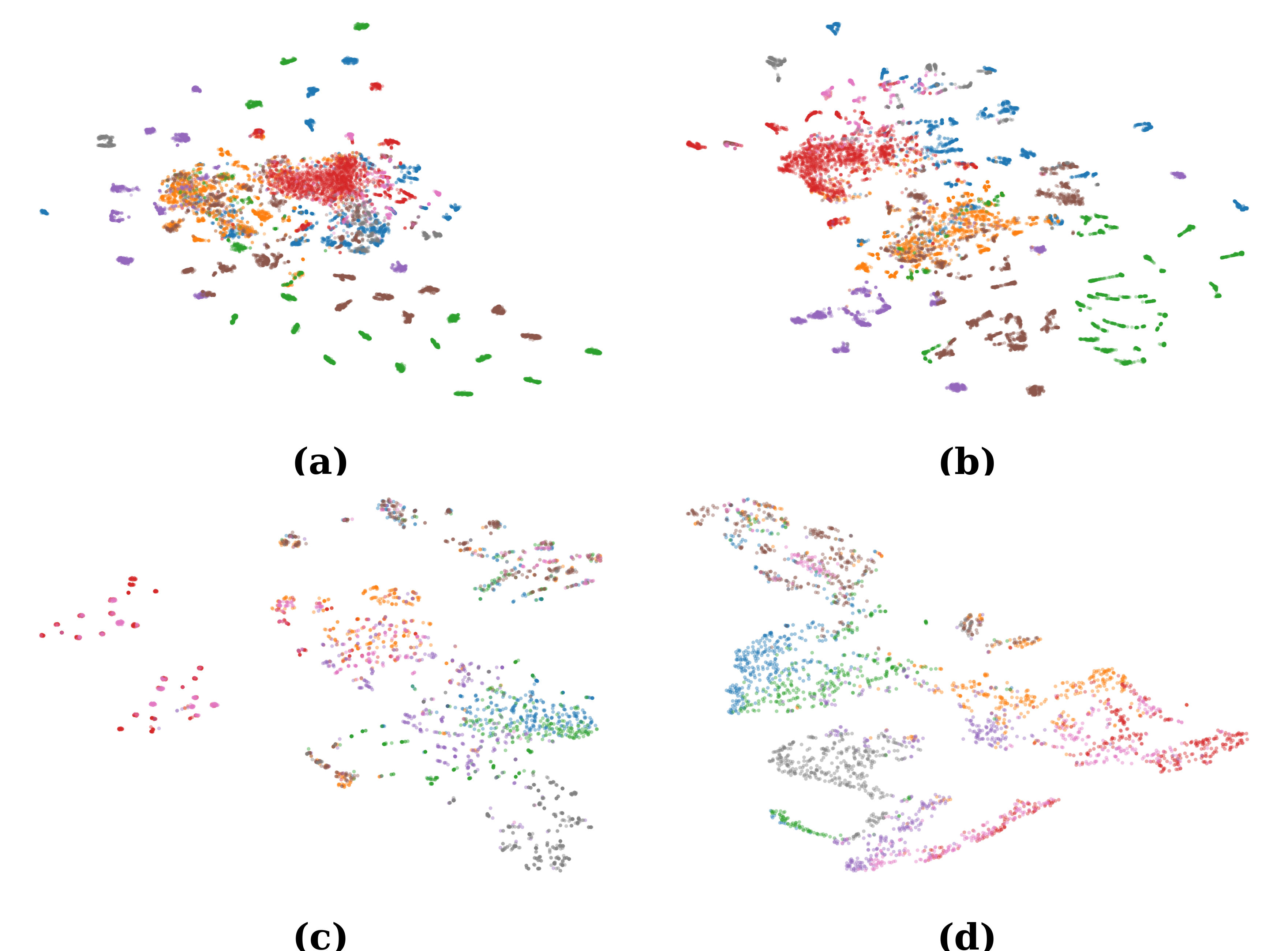}
        \captionof{figure}{
        t-SNE visualization of final-layer visual-semantic
        representations.
        Panels (a) and (c) show the base FastWAM and Cosmos Policy
        representations, while panels (b) and (d) show their
        GaussianWAM-enhanced counterparts.
        }
        \label{fig:semantic_representation}
    \end{minipage}

\end{figure*}

\paragraph{Geometric representation probing.}
We freeze the trained WAM backbones and train the same lightweight
depth probe on their final-layer visual representations, using
VGGT-Omega depth estimates as pseudo-depth supervision.
As shown in Figure~\ref{fig:depth_probe}, GaussianWAM-enhanced
representations recover clearer scene structure and more coherent depth
responses than their corresponding base models, indicating stronger
geometric grounding.

\paragraph{Visual-semantic representation visualization.}
We further visualize the learned visual-semantic representations using
t-SNE.
For each sample, the final-layer visual features are aggregated into a
feature vector and projected into two dimensions using the same t-SNE
configuration for the base and GaussianWAM models.
As shown in Figure~\ref{fig:semantic_representation}, the
GaussianWAM-enhanced representations exhibit more compact local clusters
and clearer separation among semantic groups, suggesting improved
visual-semantic organization in the learned latent space.

\section{Conclusion}
\label{sec:conclusion}

We presented \textbf{GaussianWAM}, a training-time Gaussian-field
enhancement framework for World-Action Models.
Rather than converting WAMs into explicit 3D or 4D world models,
GaussianWAM uses a 3D Gaussian field as a unified spatial teacher that
binds geometric and visual-semantic knowledge within the same 3D
representation.
The rendered semantic, depth, and alpha signals are distilled into
action-relevant WAM representations, while a validity mask restricts
supervision to reliable spatial regions.
All Gaussian-related teacher and prediction modules are removed at
inference, preserving the original WAM deployment pipeline and
computational cost.

Experiments on LIBERO, LIBERO-Plus, RoboTwin, and real-world robotic
manipulation demonstrate the effectiveness of GaussianWAM across two
distinct WAM architectures, FastWAM and Cosmos Policy.
The improvements are particularly pronounced under challenging
distribution shifts, indicating stronger geometric and visual-semantic
grounding of the learned representations.
Ablation studies further demonstrate the complementary contributions
of semantic, depth, and alpha supervision, as well as the importance
of the supervised representation depth.
Overall, our results show that training-time Gaussian distillation
provides a practical and inference-efficient approach to strengthening
the geometric and visual-semantic grounding of World-Action Models.

\clearpage
\bibliography{main}
\bibliographystyle{bibstyle}

\newpage
\setcounter{section}{0}

\end{document}